# Intracoronary Optical Coherence Tomography (OCT) Image Processing and Binary Classification Using Machine Learning

Amal Lahchim [#1], Lambros Athanasiou [#2]

[12]Electrical and Computer Engineering Department, University of Patras, Greece

[1]up1120133@upatras.gr

[2]lmathanas@uoi.gr

*Abstract*— **Intracoronary Optical Coherence Tomography (OCT) enables high-resolution visualization of coronary vessel anatomy and plaque morphology, but manual interpretation is hindered by noise, artifacts, and tissue differentiation challenges. This study proposes an automated pipeline for vessel segmentation and classification in intracoronary OCT images to reduce manual effort and computational cost. The pipeline integrates robust preprocessing, polar-to-Cartesian transformation, unsupervised K-means clustering, and local feature extraction. Logistic Regression and Support Vector Machine (SVM) models were employed for pixel-wise classification, achieving high precision (1.00) and accuracy (99.68%). The approach accurately identifies vessel boundaries with minimal false detections and outperforms traditional intensity-based methods. Future work will focus on testing with larger datasets, exploring deep learning for feature learning, and integrating real-time OCT data processing to enhance clinical outcomes.**

## I. INTRODUCTION

Invasive imaging of coronary arteries and stents deployed catheter-based using light instead of ultrasound is achieved via OCT. The principle of low-coherence interferometry is used, and therefore visualization is possible for vascular structures with an axial resolution of about 10–20 micrometers [1].

Unlike other ultrasound-based modalities, OCT obtains information about the scattering pattern of light from the tissue surface, and is therefore capable of highly precise evaluation, for example, of vessel wall architecture, fibrous cap thickness, and atheromatous plaques [2] [3].

Although the tissue penetration depth of OCT is limited to about 1 to 3 mm [1], it is sufficient to identify critical features such as calcified regions, lipid-rich cores, fibrotic tissue, and thrombus formation [2], [3], [4].

The way it works is that a light beam is shot down a fiberoptic wire, rotated, and pulled back along the artery in reverse collecting multiple axial scans (A-lines) to reconstruct high quality cross section images [5]. It yields the highest degree of imaging of microstructural changes in the coronary arteries, and provides unique views of plaque vulnerability, stent apposition, neointimal coverage and vessel healing after intervention [6].

Due to its superior spatial resolution, OCT has recently gained an unprecedented position in both research and clinical application of guiding percutaneous coronary interventions (PCI) in clinical practice [7], but this does not come without its challenges. Manual interpretation of the images is generally difficult due to the often noisy and complex images. A major source of complexity comes from the influence of lower contrast and speckle noise, as well as inhomogeneous tissue optical properties and artifacts that can blur important features and cloud interpretation [8].

In order to address these challenges, different image processing methods have been developed to increase the interpretability of OCT images. To simplify the images, thresholding techniques are used to convert grayscale intensity values to binary formats to separate tissue types [9]. Segmentation algorithms partition the images into anatomically or pathologically meaningful regions (in other words, the processes isolate Fiber walls, fibrous caps, lipid cores, or stent struts [10]. Moreover, classification of segmented regions based on machine learning and deep learning techniques further divides these regions into clinically relevant classes, such as different types of plaques and the presence of thrombus [11].

Based on this, a complete pipeline is developed for automated processing of images from intracoronary OCT focused on normal vessel anatomy in this project. Starting with preprocessing, which involves simple threshold to improve the contrast and separate the vessel lumen from the background, a coordinate transformation from Cartesian to polar coordinates is made to better accommodate the natural circular geometry of the artery. This is then followed by application of an unsupervised segmentation approach to separate vessel structures from surrounding areas without manual annotations. Textural and intensity information describing the vessel morphology in adjacent pixels is captured by a local 11×11 sliding window around each pixel to extract local features. A basic binary classification model is trained using the extracted features to classify every region as vessel or background.

## II. RELATED WORK

In the context of image analysis tasks such as segmentation, tissue classification, and vessel structure evaluation, intracoronary optical coherence tomography (OCT) has been extensively studied.

### A. Segmentation of Vessel Walls, Plaques, and Stents

OCT structure segmentation has been the focus of many works, in particular for accurate vessel wall, plaque, and stent detection. One good example is the work by Cheimariotis et al. (2017), who proposed ARC-OCT, a fully automated method to detect the lumen borders in intravascular OCT images [1]. The image transformations used by their method are based on tissue reflection and absorption, which are combined with a local regression and polynomial fitting refinement step. ARC-OCT addressed artifacts like side branches and blood presence and segmented similarly to the manual expert annotations with no significant statistical difference for most of the important geometric metrics (area, perimeter, etc.) [12].

Other researchers [13] took advantage of the need for more comprehensive vascular assessments and extended the segmentation beyond just the lumen to develop a framework for in vivo segmentation and quantification of coronary lesions from OCT images. With automated lumen segmentation, as well as identification of peculiar tissue components in the arterial wall, they proposed a method for classification of plaque type and stenosis grading. As opposed to traditional manual processes, their toolbox provides semi-automatic and offline analysis, freeing the operator as well from dependency and processing time. Furthermore, they validated their method against expert manual segmentation in 210 frames from clinical OCT data, which suggests that automatic segmentation is capable of matching human precision for lumen and plaque tissue segmentation. They also combined OCT segmentation outputs with finite element modeling to highlight the need for quantitative vessel morphology extraction for biomechanical simulations.

On this basis, further development of automatic segmentation methods has been made by Ughi et al. (2012) who developed an algorithm to quantify stent strut apposition and neointima coverage in coronary arteries on OCT imaging [14]. Their method performs the segmentation of the vessel lumen and stent struts directly from the intensity profiles along OCT A-lines. This was automated, so it could be done to evaluate stent apposition (the struts being against the vessel wall) as well as to evaluate tissue coverage (a proxy for healing). The correlation with manual expert annotations was high (Pearson's R = 0.96–0.97) and agreement in Bland–Altman analysis supported the robustness and reliability of the method. Also, many preprocessing techniques, especially some thresholding, have been deeply studied to improve structural analysis of OCT images besides segmentation.

### B. Thresholding Techniques in OCT

OCT has been pre-processed with several thresholding techniques proposed in order to do segmentation of structure. Gurmeric et al. (2009) reported one of the first methods, which involved a simple histogram-based thresholding to choose the

50th percentile of pixel intensity in splitting the vectors of bright structures (vessel wall, stent struts, and so on) from the background [15]. On the regions of interest, median and Gaussian filtering was applied to smooth the data after thresholding. Using their method, which is suited for the 3D analysis of stent strut distribution and NIH evaluation, they automatically reconstructed the lumen and stent boundaries. First, intravascular OCT images were analyzed with basic thresholding combined with targeted post-processing, and their results were shown to be in strong agreement with manual expert measurements.

Meanwhile, improvements in preprocessing steps and the combination of those early thresholding approaches with morphological operations have been adopted by more recent researchers. In fact, Pociask et al. (2018) proposed lumen segmentation from intracoronary OCT images in a fully automatic manner by combining thresholding and morphological processing [16]. The preprocessing steps in their algorithm were key and included speckle noise reduction with median and Gaussian filtering, artifact removal, and transformation from Cartesian to polar coordinates in order for the segmentation to be performed accurately. In polar space, a simple thresholding operation was then used to pick up the intimal layer, followed by morphological opening and closing of the lumen contour to remove granular artifacts. A custom linear interpolation method was implemented by the authors to address discontinuities from guidewire shadows and bifurcations. A Savitzky–Golay filter was used to smooth the contours to the final form. The method was validated on 667 OCT frames with excellent agreement with manual expert annotations (ICC = 0.97). By applying these techniques on simple thresholding based on intensity with tailored removal of artifacts and correction of contour, they showed that even simple thresholding can achieve accurate and robust lumen segmentation in complex OCT datasets.

In addition, another critical preprocessing step for OCT analysis is the transformation from polar to Cartesian coordinates.

Due to the rotating catheter within the vessel, during intravascular OCT acquisition, many raw images are acquired in polar coordinates. This makes polar to Cartesian transformation one of the common sequential preprocessing steps. Such a transformation allows the circular vessel structures to be mapped onto a rectangular Cartesian plane, so that subsequent segmentation and analysis become possible. This approach has been adopted by several studies, including Yihui Cao et al., who transformed OCT images into the polar space using a single level set method for more efficient lumen segmentation [17]. Also, other researchers like Gurmeric et al. and Pociask et al. follow the same logic by using polar transformation before thresholding and morphological operations in order to get accurate vessel structure.

In particular, Pociask et al. (2018) consider polar to Cartesian transformation to be a critical preprocessing step, to be performed before applying their fully automated lumen segmentation method in intracoronary OCT images. They converted the OCT images into polar coordinates to match the anatomic morphology of the vessel wall and removed imaging artifacts due to the presence of guidewires and circular catheter rings. The segmentation process was simplified by this transformation, which straightened the vessel cross section, giving better possibility of binarization, morphological operation, and reconstructing the lumen boundary. Finally, after segmentation in polar space, Cartesian transformation was applied to return images to Cartesian coordinates to render the final segmented lumen in its natural view. Intraclass correlation coefficient (ICC) values of 0.97 were found for the results against ground truth manual segmentations, validating their robustness and accuracy in incorporating polar to Cartesian transformation as a fundamental step for intravascular OCT image analysis [16].

Moreover, unsupervised learning techniques have been extended into selecting potentially useful training data and separating tissues.

*C. Polar-to-Cartesian Transformation*

*D. Clustering and Unsupervised Learning For Tissue Separation*

Recently, unsupervised learning methods, namely clustering, have been used in intravascular optical coherence tomography (IVOCT) image analysis in order to assist in the separation of tissues and reduction of annotation burden. An original idea of Kolluru et al. (2021) is to use deep feature-based clustering to select representative images for training segmentation models with fewer annotations [18].

Particularly, they engineered deep features learned by an autoencoder on frames of OCT, then chose diverse (based on k-medoids clustering) and informative images out of volumes of interest from these images. Traditionally, sampling strategies were equally spaced; however, this method considerably outperformed it in terms of segmentation accuracy, even using as little as 10% of all annotated data acquisition.

Feature extraction methods have also been developed to detail classification and characterization of vascular structures in OCT images that are complementary to the segmentation efforts.

*E. Feature Extraction for Classification*

Since a typical intravascular OCT image analysis often requires features extracted from local textures and pixel statistics to enforce accurate classification, meaningful information from local features is often extracted. In that paper, Tsantis et al. (2012) proposed a fully automatic segmentation framework based on the combination of local intensity distribution and wavelet-based edge information for vessel lumen border extraction and stent strut detection [19]. The Hough Transform method is first used for preprocessing images to deal with bright concentric artifacts. In this work, raw pixel intensities along with continuous wavelet transform (CWT) coefficients are combined as feature extraction of the OCT images to capture both texture and structural variations. Using this MRF framework, these features are used in an accurate lumen segmentation. The scale-space signatures extracted by Tsantis et al. were then applied for strut detection, where multi-scale wavelet analysis was used to derive scale-space signatures, feature extraction of wavelet responses (mean, variance, Lipschitz exponents, etc.), and classification on the basis of a probabilistic neural network (PNN). The results showed that combining local textural and statistical descriptors resulted in high segmentation accuracy (overlap 0.937) and strut detection performance (AUC 0.95); this demonstrates that IVOCT image classification is very sensitive to robust classification of IVOCT images.

For a more recent study, Lee et al. (2020) prototyped a hybrid feature extraction algorithm for the characterization of plaque in intravascular OCT images [20]. First, they trained a convolutional neural network (CNN) to discriminate between A-line inputs and extract 100 high-level CNN features from the activations of the CNN's fully connected layers. They simultaneously designed 371 lumen morphology features with anatomical properties, including lumen area variation, eccentricity, curvature, and border irregularities between neighboring frames. Physical phenomena in OCT imaging, such as lumen flattening close to calcifications and attenuation due to oblique beam incidence, were exploited for these morphological features. Therefore, inputs to the final feature set comprised both convolutional and morphology-based features and were fed to a random forest classifier.

Rico-Jimenez et al. (2016) also complemented these efforts by presenting a novel computational method for automatic IV-OCT plaque characterization from A-lines regarded as a linear combination of estimated depth profiles. Morphological features and LDA were subsequently used to classify these profiles into tissue types, and the profiles were derived using an alternating least squares optimization. Their method was validated against histopathology and showed that automated characterization of fibrotic and lipid-containing plaques was possible with high sensitivity and specificity, demonstrating that modeling intrinsic A-line morphological characteristics can be used to perform high-throughput and fully automated plaque characterization of intravascular OCT data [21].

III. CRITICAL ANALYSIS

Overall, despite the remarkable development in previous studies on the analysis of intracoronary OCT — segmenting vessel structures, plaque type classification, and stent apposition — most of the methods are usually complex, require extensive manual annotation, or are restricted to pathological cases. Techniques based on thresholding and polar transformations can be used to help identify basic structures but provide a significantly limited, automated apex-pipelined analysis of large scale, while lacking computational efficiency. Furthermore, current clustering and feature extraction methods are mainly dedicated to pathological tissue characterization rather than vascular anatomy characterization. For this purpose, in this work, we propose a simple and fully automated pipeline to fill the gap in the analysis of normal vessel structures from OCT images. As an approach to normal anatomy assessment in intracoronary OCT datasets, it combines basic thresholding, polar transformations, unsupervised segmentation based on a two-stage neural network, and lightweight feature-based classification.

IV. METHODS:

A. *Noise Detection and Median Filtering*

In order to enhance polar OCT image quality, three steps of noise detection and segmentation were used.

First, we estimated the Laplacian standard deviation in order to get an idea of the sharpness or clarity of the images. When the standard deviation of the Laplacian operator $\nabla_2 I$ was small, it tended to indicate that the edges were sort of smoothed perhaps due to noise.

Second, we considered salt-and-pepper noise, modeling it as the ratio between extreme pixel values:

$$\text{SPR} = \frac{|\{x \in I \mid x \leq \tau_{\min} \text{ or } x \geq \tau_{\max}\}|}{N}$$ [24]

where $\tau_{\min}=5$, $\tau_{\max}=250$, and $N$ is the total number of pixels. The value SPR > 0.75 was set to define substantial contamination.

An 11 by 11 sliding window was used in calculating local variance to assist in identifying speckle noise. In regions that were homogeneous, high local variance tended to indicate multiplicative noise.

$$I'(x,y) = \text{median}\{I(i,j) \mid (i,j) \in \mathcal{N}(x,y)\}$$ [24]

A lot of frames contained a high salt-and-pepper ratio, which is why we used a median filter with k =3.

This is a nonlinear filter which was able to suppress impulse noise and keep structural edges.

B. *Guidewire Shadow Removal*

We eliminated the guidewire artifact by selecting the vertical area of least intensity.

The vessel wall was moved back onto this region with a shift-and-blend technique and cropped.

In order to form a smooth transition and low visual interruption along the cut seam, a blending region was formed in the cut seam.

The feature derived from the blended vessel image was not involved in training or visualization of the blended vessel image.

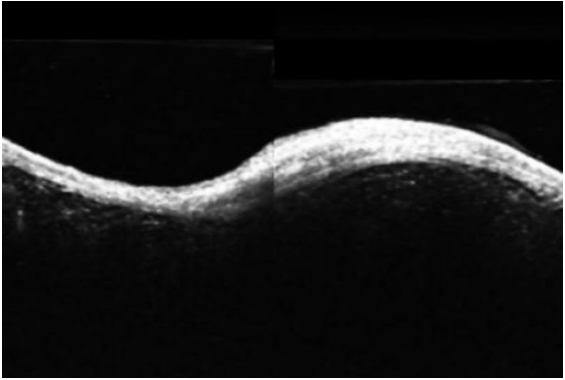

**Fig1. Blended OCT Image After Guidewire Removal**

C. *Otsu's Thresholding*

Otsu's method was implemented for this project – a common automatic thresholding method.
The Otsu's method selects the threshold such that the intra-class variance of the thresholded black and white image is minimized, or equivalently, so that inter-class variance is maximized.

This approach is well adapted when the grayscale histograms are bimodal, such as generally obtained in medical imagery, where two distinct peaks are usually corresponding to tissue and non-tissue regions.

The Otsu's method is data-dependent; the threshold is adjusted according to the intensity characteristics of the image when this method is used.

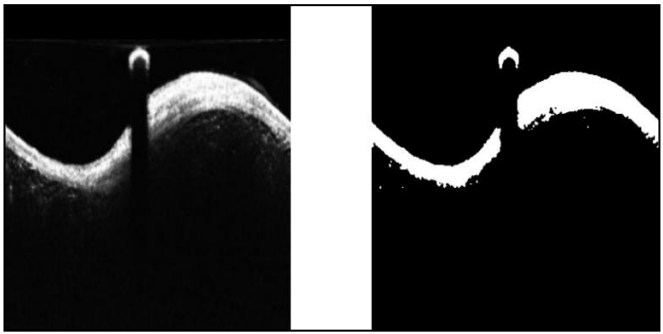

**Fig2. Original OCT image showing a prominent guidewire artifact (left) and the corresponding binary mask highlighting vessel wall structures while preserving the artifact region (right)**

D. *Polar to Cartesian Transformation*

Once we have filtered the image, we then use opencv's cv2.remap() to map the image from polar to Cartesian coordinates and process it with common image analysis techniques.

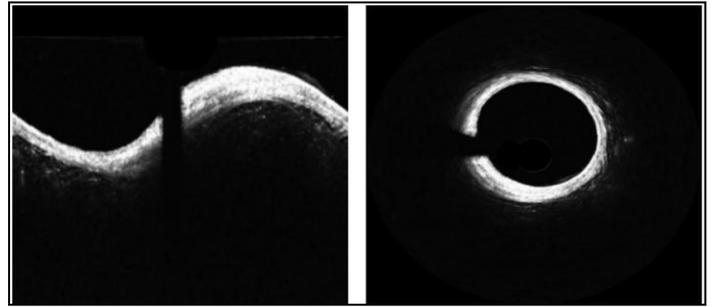

**Fig3. ORGINAL OCT IMAGE DISPLAYED Both in polar(left) and cartesian (right)**

E. *K-Means-Based Vessel Segmentation with Morphological Refinement in Cartesian OCT Images:*

K-means clustering was applied to the grayscale, cartesian-transformed OCT image to perform unsupervised segmentation of the vessel lumen and surrounding tissue, the image was first reshaped into a 1D array of intensity values to be used as input for the K-means algorithm (k = 2 clusters, vessel vs. background), then Cluster centroids were updated.

At each iteration as the mean intensity of the assigned pixels, and convergence was checked based on minimal centroid movement, and the intermediate outputs from each iteration were captured and displayed as subplots to illustrate the segmentation process.

these visuals in figure 4 is providing a clear view of how the vessel regions are progressively separated from the background based on intensity differences.

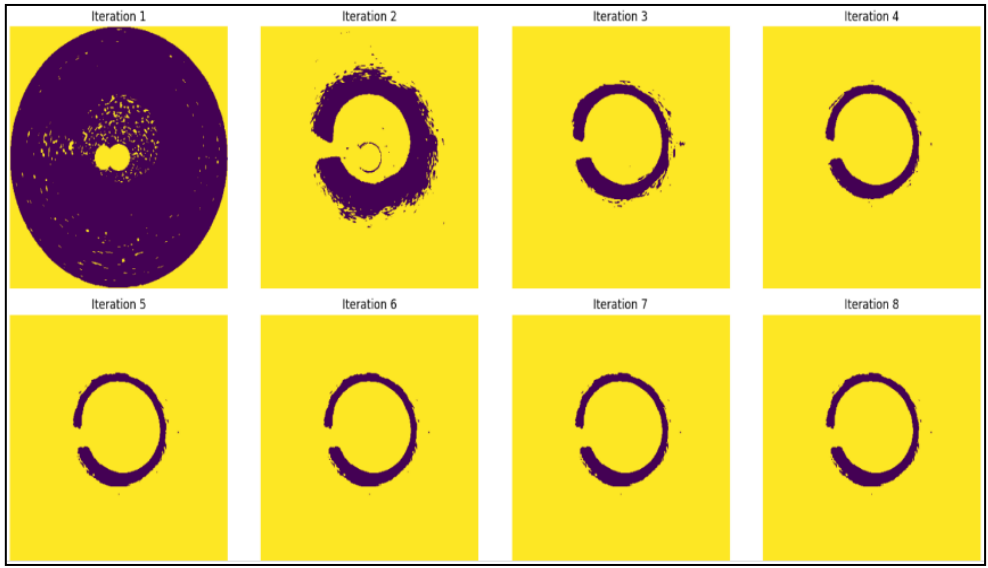

**Fig4. Progressive Vessel Segmentation Using K-means Clustering**

The inertia (the sum of squared distances of samples to their closest cluster center) was calculated after each iteration to assess the convergence of the K-means algorithm.

$$\text{Inertia} = \sum_{i=1}^{k} \sum_{x \in C_i} \|x - \mu_i\|^2 \quad [25]$$

In this equation, k is the total number of clusters, $C_i$ denotes the set of points in cluster i, x is a data point belonging to $C_i$, and $\mu_i$ is the centroid of cluster i. The term $\|x-\mu_i\|^2$ represents the squared Euclidean distance between the data point and its corresponding cluster center.

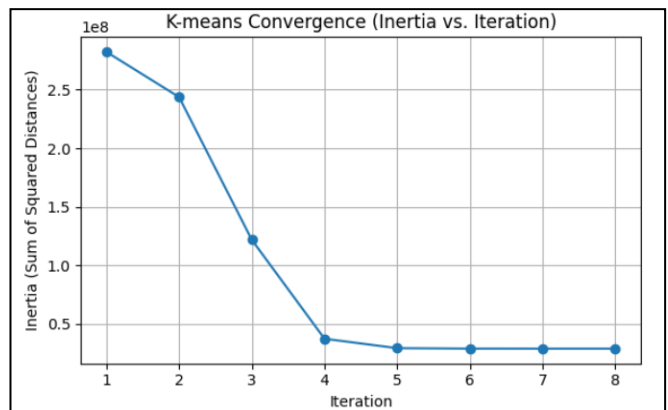

**Fig5. K-means convergence**

Figure 5 indicates the inertia is largely inhibited in the initial iterations; the inertia stopped declining and This validates that, with this sudden drop, the clusters started to separate well in the initial iterations and then stabilized.
Moreover, the plot shows that within a maximum of 5–6 stabilized at the collision number 4, implying the clusters reached a stable configuration quickly. iterations, the algorithm converged to a stable segmentation result. This observed behavior aligns with the low-dimensional intensity data that was defined a priori for this problem.

F. *Feature Extraction*

Local descriptors for classification were sampled around each pixel using a sliding window method with a patch size of 11 × 11 on the Cartesian OCT image.

Since the filter's sliding window must be centered around each pixel (otherwise, pixels on the edges would not have sufficient surrounding context), both the grayscale image and corresponding binary mask were symmetrically padded with the help of OpenCV's copyMakeBorder with cv2.BORDER_REFLECT.

The padding size was half of the patch width so that the padding covers half of the patch width beyond the patch boundary on each side of the patch border and the complete image is covered, but no relevant information is lost near the border.

A square patch was cropped centered at each pixel from the padded image. The seven features were calculated for each patch to represent both intensity and texture information:

• Mean intensity: the average pixel value, reflecting local brightness.

• Standard deviation: estimate contrast or local variation.
• Minimum and maximum intensity: local intensity range captured
• Median intensity: less sensitive to outliers
• Entropy: calculated with a 3×3 disk kernel using local texture complexity value
• Magnitude of gradient: determined by the summation of Sobel-x and Sobel-y filters, characterizing the magnitude of edges in the patch.

The padded segmentation mask (equally padded) served for the labelling of each pixel with a class. If the corresponding value of the pixel in the mask is 255, it was labeled as vessel (class 1), otherwise, it was labeled as background (class 0).

This mask had the effect that feature extraction could produce a supervised training set, in which individual feature vectors were registered in a prior unsupervised segmentation to anatomical structure.

**Since OCT segmentation mask has natural class imbalance, background samples were downsampled randomly to be equal to the amount of vessel samples. This well-balanced dataset is useful in training a classifier that is not biased on the majority class Segmentation accuracy.**

G. *Feature Analysis and visualization for vessel classification*

HISTOGRAM DISTRIBUTION OF EXTRACTED FEATURES:

The distribution histogram for each of the seven features mean, standard deviation, minimum, maximum, median intensity, entropy, and Sobel gradient magnitude—were plotted to evaluate the capability of each feature in discriminating the pixels belonging to the vessel and those on the background.

Using the balanced dataset, we visually observed the histograms of vessel and background pixels independently.

According to the results, average intensity, entropy, and gradient magnitude, among other characteristics, showed separable features.
The pixels of vessels had higher entropy and gradient due to the complicated structure and the active edges of vessels, while the pixels of background were more homogeneous, with lower variance and entropy.

On the other hand, other features, like the ones related to minimum intensity, presented greater interclass overlapping, not being so contributive to classification.

In this analysis, the feature with the lowest intensity level was eliminated from the feature set, leading the process of feature selection, and also indicating which descriptor is informative.

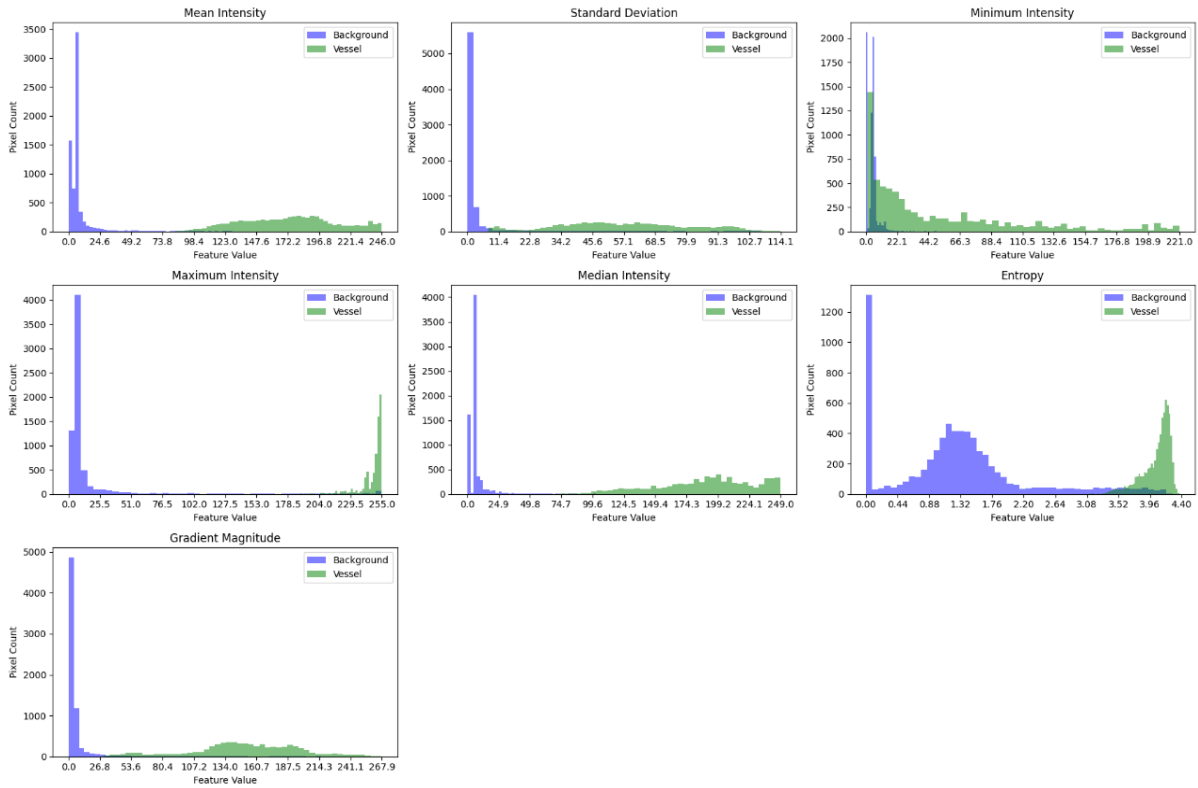

*Fig5.Distribution Analysis of Extracted Features for Vessel and Background Pixels in OCT Images*

H. *Feature-wise PCA Visualization of Vessel vs Background Data*

In order to better visualize the data and how the features are distributed in it, as well as understanding how they behave, the dimension of the 7-dimensional feature space was reduced down to 2 dimensions using PCA.

This allowed us to incorporate this information in attempting to visualize the spatial placement of samples and in assessing class separability in a lower-dimensional space.

The figure 6 shows each subplot for a 2D PCA projection, where points are colored based on one of the original features (mean intensity, entropy, gradient magnitude, etc.)

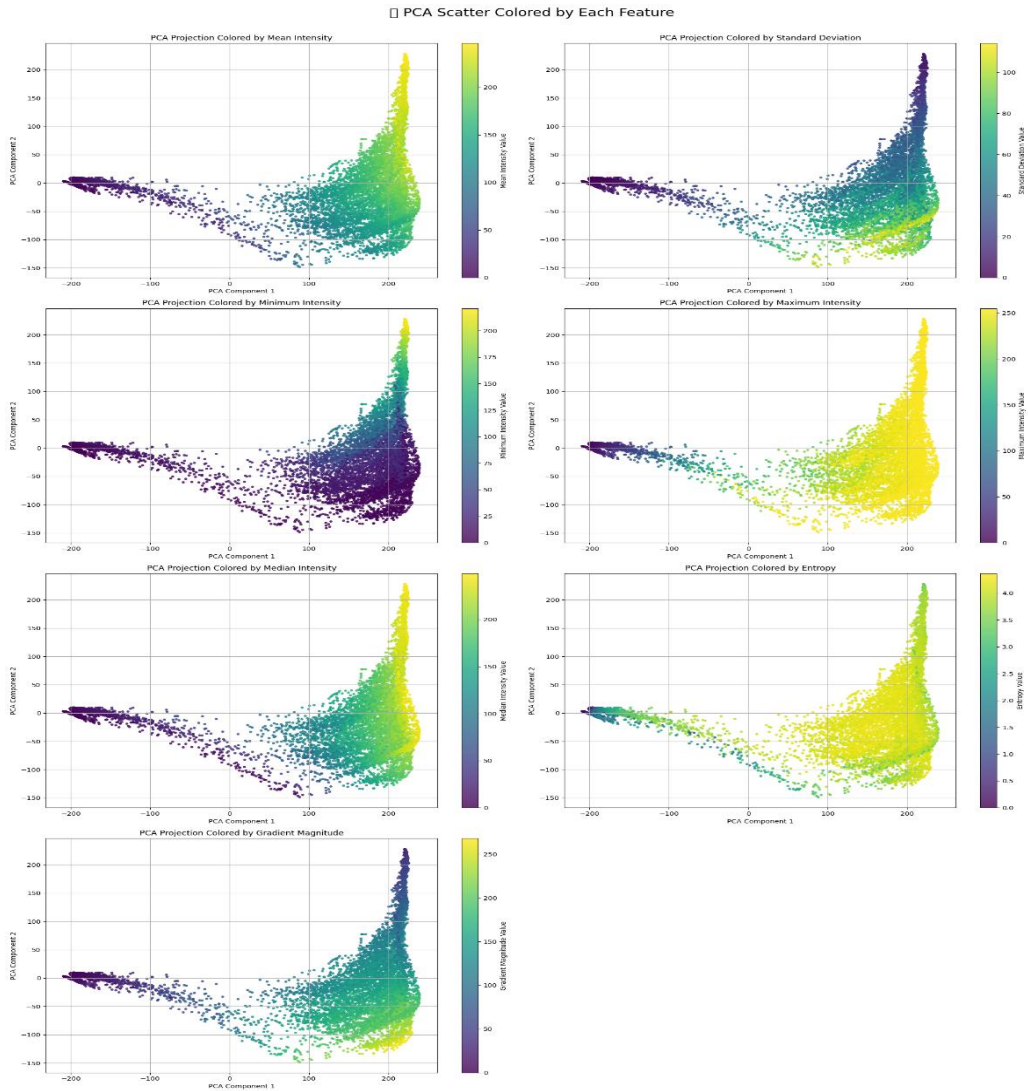

*Fig 6. Two-Dimensional PCA Projection Colored by Individual Feature Values*

## V. RESULTS AND DISCUSSION

Machine learning models evaluation: A balanced dataset made up of seven extracted features per pixel was evaluated to classify vessel pixels in OCT images using 2 machine learning models. The Logistic Regression model provided almost perfect classification as traits (precision, recall, and F1-score) were all 1.00 for both the vessel (class 1) and the background (class 0). This robustness was seen in the confusion matrix that revealed with 3 false negatives and 5 false positives among a total of 2,782 test pixels.

Similarly, the Support Vector Machine (SVM) classifier having a RBF kernel fit too had superb performance. On background pixels, the precision, recall, and F1-score had respective values of 1.00, 0.99, and 1.00; on vessel pixels, the values of the three measures were 1.00. The respective confusion matrix showed 7 background pixels incorrectly categorized as vessels and also showed 2 vessel pixels incorrectly categorized as background, which showed a total of 9 classification errors, where the overall accuracy provided was 99.68%.

The trained SVM was further checked using its spatial generalization property by executing pixel-wise segmentation on the complete OCT image. The output of this prediction mask, which was resized back to the normal dimensions of the image, was naturally compared with the original ground truth mask, which was derived with the help of K-means clustering. As shown in Fig 7, it can be seen that the anatomical vessel boundary representation by the SVM is very close to the expected shape. Besides, the obtained SVM segmentation over the usual Cartesian OCT image Fig. 8 reveals the correct localization of the inner vessel wall.

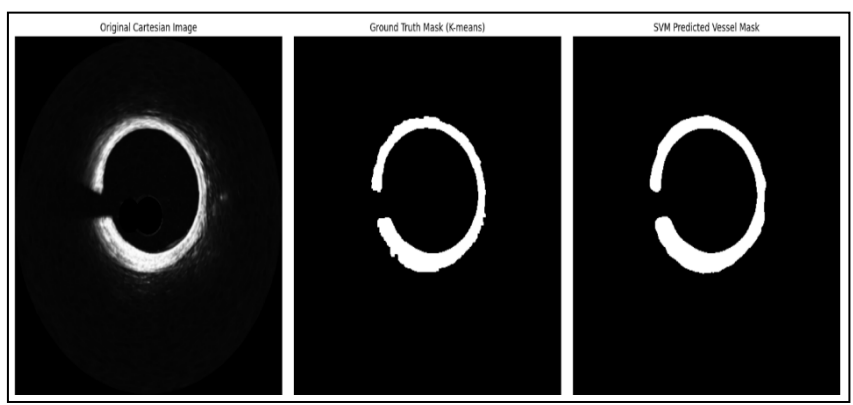

*Fig 7. Comparison of K-means and SVM-Based Vessel Segmentation on OCT Image*

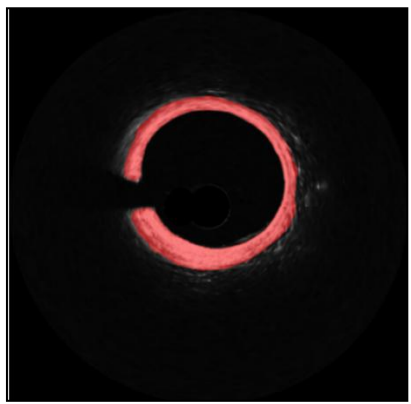

*Fig 8. Overlay of SVM-Predicted Vessel Segmentation on OCT Image*

The comparative presentation of techniques with the actual practices is provided in Table 1. This underlines the strengths of the enhanced accuracy and soundness of the suggested supervised learning technique in contrast to the commonplace intensity-based approaches or unsupervised learning strategies.

*Table 1: Comparison of Intracoronary OCT Image Segmentation and Classification Approaches*

| STUDY / METHOD | INPUT | TECHNIQUE | ACCURACY / DICE / OTHER | NOTES | REFERENCES |
|---|---|---|---|---|---|
| MORAES ET AL. (2013) | IVOCT IMAGES | WAVELET TRANSFORM + OTSU THRESHOLD + BINARY MORPHOLOGICAL RECONSTRUCTION | TP = 99.29% FP = 3.69% ± 2.88% FN = 0.71% ± 2.96% OVERLAP DICE ≈ 98% | FULLY AUTOMATIC, TESTED ON 209 IMAGES (HUMAN, PIG, RABBIT CORONARIES); ROBUST FOR CHALLENGING LUMEN SHAPES | [21] |
| FUJINO ET AL., 2018 | IVOCT PULLBACKS (PRE-STENTING) | MANUAL CALCIUM SCORING (ANGLE, THICKNESS, LENGTH) | AUC 0.86 FOR STENT UNDER-EXPANSION PREDICTION | CALCIUM SCORE: ANGLE >180° = 2 PTS, THICKNESS >0.5 MM = 1 PT, LENGTH >5 MM = 1 PT; OUTPERFORMED ANGIOGRAPHY | [22] |
| KOCH ET AL., 2025 | OCT PULLBACKS (1148 FRAMES, 92 PULLBACKS HUMAN; RABBIT MODEL FOR VALIDATION) | DEEP LEARNING: UNET++ FOR SEGMENTATION, RESNET-18 FOR QUADRANT-LEVEL CLASSIFICATION | SEGMENTATION DICE: LUMEN 0.99, STENT 0.66, NEOINTIMA 0.86; CLASSIFICATION ACCURACY: 75% (HUMAN TEST SET), 87% (ANIMAL) | FULLY AUTOMATED NEOINTIMA CHARACTERIZATION; COMPARABLE TO EXPERT MANUAL ANALYSIS; OPEN-ACCESS TOOL; WELL-CALIBRATED CONFIDENCE SCORES | [23] |
| CURRENT WORK | INTRACORONARY OCT IMAGES (POLAR & CARTESIAN) | PREPROCESSING, POLAR–CARTESIAN TRANSFORM, K-MEANS SEGMENTATION, FEATURE EXTRACTION, LOGISTIC REGRESSION & SVM CLASSIFICATION. | LOGISTIC REGRESSION PRECISION/RECALL/F1 = 1.00; SVM ACCURACY 99.68% | FOCUS ON NORMAL ANATOMY; MINIMAL MANUAL ANNOTATION; ROBUST PIXEL-WISE VESSEL VS BACKGROUND CLASSIFICATION | |

As shown in Table 1, the proposed study consists of a simple but efficient pipeline, which might be more accurate than other unsupervised methods since most intensity-based techniques and comparable to more complex deep learning models, although it does not significantly require any computational resources or rely on manual annotation to train. This is a potentially feasible method in both ways of carrying out large-scale anatomical analysis in a more cost-effective and less sophisticated way, which will help it to adopt such a method in general clinical practice and interpretation of results of OCT images in real-time.

## VI. Conclusion

The current work is demonstrating that, despite using a simple yet extremely effective pipeline, it becomes obvious that high-accuracy vessel segmentation and their classification in images obtained by intracoronary OCT can be performed without the extensive use of computing resources and manual labeling. The approach presents high-quality vessel boundary detection that is even superior to a good deal of the historic approaches that used intensity-based detection, within an impressively lightweight frame.

The further area of strength and clinical significance of this method is predetermined by the successful results of Logistic Regression and SVM in the future. It is hoped that in the future the pipeline will be tested on bigger, more varied datasets, including difficult cases and a variety of imaging conditions. Moreover, the automatic learning of the features with the help of deep learning structures, a study of temporal regularity of the 3D OCT series, as well as upgrading post-processing based on the application of advanced methods can also help bring the quality of the segmentation to an even higher level. Lastly, the combination of the pipeline with real-time OCT analysis devices will enable a more assured and faster evaluation of the vessels in the course of the intervention, and this means that it can be used in a wide-scale clinical setting and will yield satisfactory patient results.